\documentclass[letterpaper, 10 pt, conference]{ieeeconf}  

\IEEEoverridecommandlockouts                              

\usepackage{soul}
\usepackage{booktabs}
\usepackage{subfig}
\usepackage{amsmath} 
\usepackage{amssymb}  

\usepackage{float}
\usepackage{graphics} 
\usepackage{graphicx}
\usepackage{multirow}
\usepackage{indentfirst}
\usepackage{makecell}
\usepackage{wrapfig}
\usepackage{mathrsfs}
\usepackage{amsfonts}
\usepackage{kantlipsum}

\makeatletter
\let\NAT@parse\undefined
\makeatother
\usepackage[numbers,sort&compress]{natbib}
\usepackage[pagebackref,breaklinks,colorlinks=true,citecolor=green]{hyperref}

\title{\LARGE \bf
DeformPAM: Data-Efficient Learning for Long-horizon \ul{Deform}able Object Manipulation via \ul{P}reference-based \ul{A}ction Align\ul{m}ent}

\author{Wendi Chen$^{12*}$, Han Xue$^{12*}$, Fangyuan Zhou$^{1}$, Yuan Fang$^{1}$ and Cewu Lu$^{13}$
\thanks{$^{1}$Shanghai Jiao Tong University. $^{2}$Meta Robotics Institute, SJTU. $^{3}$Shanghai Innovation Institute. 
$^*$ indicates equal contribution.
{\tt\small\{chenwendi-andy, xiaoxiaoxh, ui-micro, sjtu\_fy, lucewu\}@sjtu.edu.cn}}%
}%
\begin{document}
\newcommand{\framework}{DeformPAM}
\newcommand{\etc}{\textit{etc}}
\newcommand{\etal}{\textit{et al.} }
\newcommand{\ie}{\textit{i}.\textit{e}., }
\newcommand{\eg}{\textit{e}.\textit{g}., }

\makeatletter
\let\@oldmaketitle\@maketitle
\renewcommand{\@maketitle}{
\@oldmaketitle
}
\makeatother

\maketitle
\thispagestyle{empty}
\pagestyle{empty}

\begin{abstract}
In recent years, imitation learning has made progress in the field of robotic manipulation. However, it still faces challenges when addressing complex long-horizon tasks with deformable objects, such as high-dimensional state spaces, complex dynamics, and multimodal action distributions. Traditional imitation learning methods often require a large amount of data and encounter distributional shifts and accumulative errors in these tasks. To address these issues, we propose a data-efficient general learning framework (\framework) based on preference learning and reward-guided action selection. \framework~decomposes long-horizon tasks into multiple action primitives, utilizes 3D point cloud inputs and diffusion models to model action distributions, and trains an implicit reward model using human preference data. During the inference phase, the reward model scores multiple candidate actions, selecting the optimal action for execution, thereby reducing the occurrence of anomalous actions and improving task completion quality. Experiments conducted on three challenging real-world long-horizon deformable object manipulation tasks demonstrate the effectiveness of this method. Results show that \framework~ improves both task completion quality and efficiency compared to baseline methods even with limited data. Code and data will be available at \href{https://deform-pam.robotflow.ai}{deform-pam.robotflow.ai}.
\end{abstract}

\vspace{-0.2cm}
\section{Introduction}
\vspace{-0.2cm}
Efficiently learning to perform general manipulation tasks has been a persistent focus in the field of robotics. In recent years, imitation learning has made significant progress in robotic manipulation \cite{chi2023diffusionpolicy, ACT, Ze2024DP3, chi2024universal, aloha_unleashed}. However, these algorithms usually require a huge amount of data (\eg thousands of demonstrations) for complex long-horizon deformable object manipulation tasks \cite{aloha_unleashed}. Such tasks present the following unique properties: 
\begin{itemize}
    \item \textbf{High-dimensional state space} that often leads to complex initial and intermediate object states.
    \item \textbf{Complex dynamics} that are difficult to accurately model in simulations.
    \item \textbf{Multi-modal distribution} in action space.
\end{itemize}
\begin{figure}[htbp]
  \centering
  \includegraphics[width=0.48\textwidth]{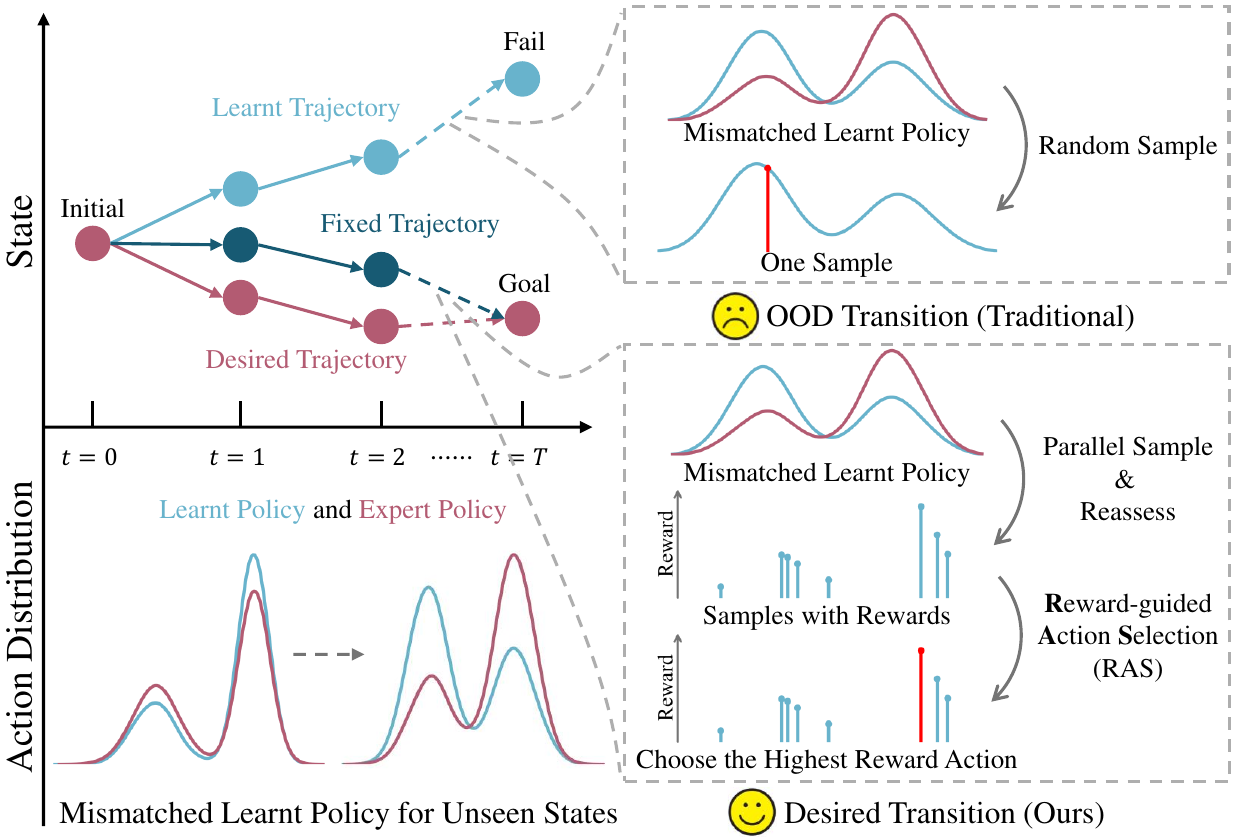}
  \caption{In long-horizon manipulation tasks, a probabilistic policy may encounter distribution shifts when imperfect policy fitting leads to unseen states. As time progresses, the deviation from the expert policy becomes more significant. Our framework employs \textbf{R}eward-guided \textbf{A}ction \textbf{S}election (\textbf{RAS}) to reassess sampled actions from the generative policy model, thereby improving overall performance.}
  \label{fig:teaser}
  \vspace{-0.5cm}
\end{figure}
These characteristics lead to significant distribution shifts and accumulation errors in traditional imitation learning algorithms for complex long-horizon tasks \cite{Dagger}. For an action policy modeled with probabilistic models (e.g., diffusion \cite{chi2023diffusionpolicy}), encountering unseen complex states makes the agent gradually drift away from the desired trajectory (see Fig. \ref{fig:teaser} left). To ensure that the data covers the high-dimensional state space as much as possible, the cost of collecting data in the real world will increase significantly. So, how can we learn to perform complex long-horizon deformable object manipulation tasks with a limited amount of data?

Our core idea is simple: we try to make the policy model distinguish between \textit{good} and \textit{bad} actions and only select the best action (see Fig. \ref{fig:teaser} right) during inference. This will reduce abnormal actions and alleviate distribution shifts in long-horizon tasks. Reward functions are a common way to evaluate actions, but as in previous works  ~\cite{clothfunnels, avigal2022speedfolding}, designing a reward function for each task individually has much hidden cost. Therefore, we choose to use human preference data as a general representation across tasks for evaluating action quality.  

Based on this idea, we propose a general learning framework \textbf{\framework} (see Fig. \ref{fig:pipeline}) for long-horizon \textbf{Deform}able object manipulation via \textbf{P}reference-based \textbf{A}ction align\textbf{M}ent. Our approach has three stages: \textbf{(1)} In the \textbf{first stage}, we collect a small amount of human demonstration data and use supervised learning to train an initial probabilistic policy model based on diffusion \cite{DDPM} and action primitives. \textbf{(2)} In the \textbf{second stage}, we run rollouts on real robots with the initial probabilistic policy model and record the $N$ predicted actions for each state, which are then annotated with preference data by humans. We use DPO (Direct Preference Optimization) \cite{DPO} on diffusion models \cite{diffusion_dpo} to directly learn an implicit reward model from this preference data. \textbf{(3)} Finally, during \textbf{inference}, we use \textbf{R}eward-guided \textbf{A}ction \textbf{S}election (\textbf{RAS}) to boost the performance of the initial policy model from the first stage. Specifically, we use the initial policy model to generate $N$ actions, score them using the implicit reward model, and select the action with the highest reward score to execute.

We find that this approach effectively reduces the occurrence of anomalous actions, thereby improving the performance of complex long-horizon tasks for deformable objects.

To validate the effectiveness of our learning framework, we conducted extensive real-world experiments on three challenging long-horizon deformable object manipulation tasks involving granular (granular pile shaping), 1D (rope shaping), and 2D (T-shirt unfolding) deformable objects. All these tasks start with very complex initial object states. We use IoU, coverage, and Earth Mover's Distance (EMD) to quantitatively measure the task completion quality of the model. Real-world results indicate that our method improves task completion quality and time compared to baseline methods across multiple complex tasks. Our contributions are summarized as follows:
\begin{itemize}
    \item We design a general primitive learning framework (\framework) for long-horizon deformable object manipulation, which uses an implicit reward model trained by preference data to select the action with higher quality.
    \item We evaluate our method with real robots on several highly challenging long-horizon deformable object manipulation tasks.
\end{itemize}


\section{Related Works}
\subsection{Deformable Object Manipulation}
Deformable object manipulation is a field with a long research history and numerous applications. Most methods in this domain typically construct specific simulation environments tailored to particular object types \cite{unifolding, ha2022flingbot, wu2024unigarmentmanip, xu2023roboninja}, designing specialized rewards \cite{clothfunnels, avigal2022speedfolding} or learning pipelines \cite{Wang2023One, chen2023autobag} to accomplish specific tasks. These hidden costs make it challenging for these learning frameworks to generalize across tasks. Recently, Differentiable Particles \cite{chen2024differentiable} attempted to use a differentiable simulator to plan optimal action trajectories applicable to various tasks. However, it requires additional object state estimators as input, whereas our approach learns actions directly from raw point clouds. AdaptiveGraph \cite{zhang2024adaptigraph} is a model-based method for general-purpose deformable object manipulation, which learns the dynamics model of deformable objects using massive data in simulation and online interaction data in the real world, followed by using MPC to plan optimal execution trajectories. However, like Differentiable Particles \cite{chen2024differentiable}, this method requires building simulation environments for each object type and each task, and it also suffers from the sim-to-real gap due to complex dynamics of deformable objects.

\subsection{Imitation Learning for Long-horizon Manipulation}
In recent years, there have been two main approaches to extend imitation learning to complex long horizon tasks: hierarchical imitation learning~\cite{mandlekar2020iris,mandlekar2020learning,shiarlis2018taco, xu2018neural, gao2024prime} and learning from play data~\cite{wang2023mimicplay, lynch2020learning, cuiplay, rosete2023latent}. Hierarchical imitation learning decomposes task learning into high-level planning and low-level controllers, while the latter approach collects interaction environment data through human teleoperation of robotic arms, without requiring specific task goals. Our method is more akin to hierarchical imitation learning, which improves sample efficiency by utilizing atomic action skills. However, these learning methods usually perform experiments on long-horizon tasks with rigid objects~\cite{gao2024prime, rosete2023latent}, or assume simple initial object states~\cite{wang2023mimicplay} (\eg flattened cloth). In comparison, our framework focuses on long-horizon tasks with deformable objects in complex initial states. RoboCook \cite{shi2023robocook} is a framework for learning long horizon tasks involving deformable objects, but it is specifically designed for elasto-plastic objects (\ie dough), making it difficult to adapt directly to 1D (e.g., ropes) and 2D (e.g., garments) deformable objects. In contrast, our method theoretically applies to deformable objects of various dimensions.

\subsection{Learning from Human Preference}
Learning from human preference data \cite{sadigh2017active, ibarz2018reward, biyik2020active} has garnered attention in the field of robotics.
Recently, reinforcement learning from human feedback (RLHF) \cite{christiano2017deep,instruct_gpt} has become a popular way of leveraging preference data for aligning policy models (\eg large language models).
Subsequently, to eliminate the reliance on an explicit reward model in RLHF, DPO~\cite{DPO} and CPL~\cite{CPL} enable direct policy finetuning from preference data, based on contextual bandits and Markov decision processes respectively.
Additionally, PFM~\cite{kim2024preference} learns a conditional flow matching model from preference data to optimize the actions predicted by the policy model.
Owing to their convenience, this methodology has also been applied in fields like image generation (Diffusion-DPO~\cite{diffusion_dpo}).
Instead of directly using the finetuned policy model~\cite{diffusion_dpo} or learning an action transformation model~\cite{kim2024preference}, we leverage the underlying implicit reward model of DPO to guide action selection from multiple generated action samples, which has been proven to be beneficial in natural language processing (NLP) \cite{askell2021general,khanovargs}.


\section{Preliminary}
\subsection{Conditional Diffusion Models}
Diffusion models are a series of generative models that excel at generating samples $x_0$ from arbitrary multimodal distributions by progressively denoising Gaussian noise $x_T$.
They can be conditional when given some condition $c$.
A conditional diffusion model comprises two processes: the forward diffusion process and the reverse denoising process. They are considered as a Markov chain with fixed transitions $q$ and learnable transitions $p_\theta$ respectively, which can be expressed as
\vspace{-0.2cm}
{\small
\begin{align}
    q(x_t|x_{t-1}) &: x_t = \alpha_t^{1/2}x_{t-1} + (1-\alpha_t)^{1/2}\epsilon_{t-1}, \\
    p_\theta(x_{t-1}|x_t, c) &: x_{t-1} = \mu_\theta(x_t, c, t)+(\Sigma_\theta(x_t, c, t))^{1/2}\xi_{t-1}.
    \vspace{-0.2cm}
\end{align}
}
where $\{\alpha_t \in (0, 1)\}_1^T$ are the predefined variance schedule and $\epsilon, \xi$ are Gaussian noise.  Moreover, an expression for directly calculating the diffusion result can be written as
\vspace{-0.2cm}
\begin{equation}
    \label{eq:direct_diffusion}
    q(x_t|x_0): x_t = \prod_{i=1}^t \alpha_i^{1/2}x_0+(1 - \prod_{i=1}^t \alpha_i)^{1/2}\epsilon.
    \vspace{-0.2cm}
\end{equation}
During training, reparameterize $\mu_\theta$ as $\mu_\theta(\epsilon_\theta,x_0)$ with Eq.~\ref{eq:direct_diffusion} and a simplified ELBO objective in DDPM~\cite{DDPM} is derived as
\begin{equation}
    \vspace{-0.2cm}
    L_{simple} = \mathbb{E}_{x_0, t, \epsilon}\|\epsilon - \epsilon_\theta(x_t, c, t)\|_2^2.
    \vspace{-0.2cm}
\end{equation}


\begin{figure*}[h]
  \centering
  \includegraphics[width=1.0\textwidth]{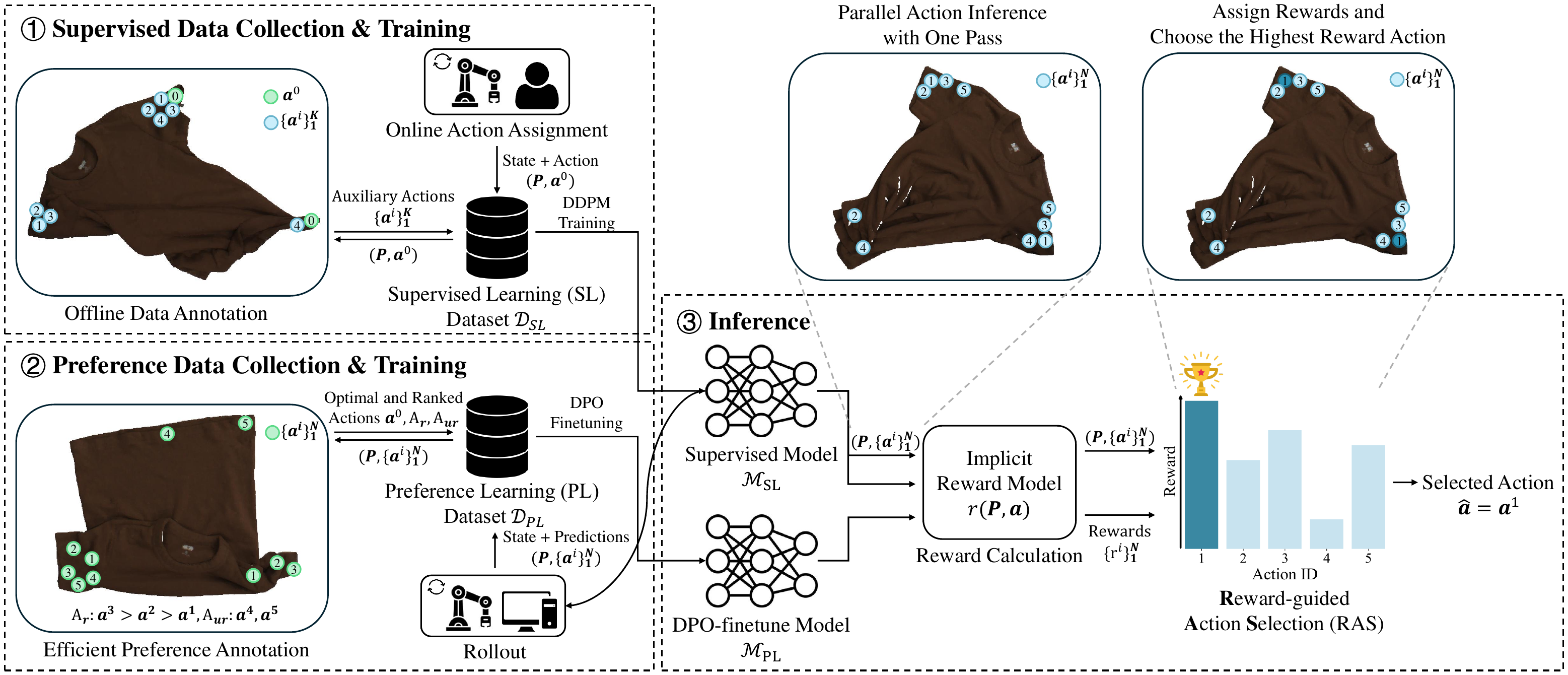}
  \caption{Pipeline overview of \textbf{\framework}. (1) In \textbf{stage 1}, we assign actions for execution and annotate auxiliary actions for supervised learning in a real-world environment and train a supervised primitive model based on Diffusion. Circles with the same numbers represent the manipulation positions for an action. (2) In \textbf{stage 2}, we deploy this model in the environment to collect preference data composed of annotated and predicted actions. These data are used to train a DPO-finetuned model. (3) During \textbf{inference}, we utilize the supervised model to predict actions and employ an implicit reward model derived from two models for \textbf{R}eward-guided \textbf{A}ction \textbf{S}election (\textbf{RAS}). The action with the highest reward is regarded as the final prediction.}
  \label{fig:pipeline}
  \vspace{-0.5cm}
\end{figure*}

\section{Methodology}
\vspace{-0.2cm}
We will illustrate our learning pipeline \textbf{\framework}  (see Fig. \ref{fig:pipeline}) in three parts: (1) In Section \ref{sec:sl}, we demonstrate how to train a diffusion-based primitive policy model with supervised learning. (2) In Section \ref{sec:pl}, we describe how to use preference data to learn an implicit reward model with DPO finetuning. (3) In Section \ref{sec:ras}, we present \textbf{R}eward-guided \textbf{A}ction \textbf{S}election (RAS), which boosts the performance of the supervised model during inference by using an implicit reward model to guide action selection.

\subsection{Supervised Learning for an Initial Primitive Policy}
\vspace{-0.15cm}
\label{sec:sl}
We will first introduce the basics of action primitive learning in Sec. \ref{sec:primitive_learing}. Then we will illustrate how to collect data to train an initial primitive policy model with supervised learning in Sec. \ref{sec:sl_data_collection} and Sec. \ref{sec:sl_network}.
\subsubsection{Action Primitive Learning}\label{sec:primitive_learing}
In order to improve data efficiency, we decompose long-horizon tasks into multiple action primitives, and our model predicts only the action parameters for each primitive. This approach not only reduces the horizon length \cite{gao2024prime} but also allows us to perform highly dynamic actions (e.g. fling a garment \cite{ha2022flingbot}). Our primitive learning network takes an RGB-D image $\mathcal{I}$ as input in each manipulation step. After that, Grounded SAM~\cite{ren2024grounded} is used to segment the point cloud $\mathbf{P}_t$ of the target object, then the policy model $\mathcal{M}$ will predict the predefined primitive action $\hat{\mathbf{a}} = \mathcal{M}(\mathbf{P})$. We use OMPL~\cite{sucan2012open} to generate planning trajectories based on primitive parameters. 
PyBullet\cite{coumans2016pybullet} and rule-based criteria are employed to ensure safety.
\subsubsection{Data Collection}\label{sec:sl_data_collection}
In order to acquire demonstration data, we design a graphic interface to let the user annotate one optimal action primitive parameter $\mathbf{a}^0$ for each observation step and let the robot execute the action primitive. Since the optimal actions in deformable objects manipulation are often diverse (multi-modal), we offline annotate additional $K$ potentially optimal actions $\{\mathbf{a}^k\}_1^K$ called auxiliary actions (see Fig. \ref{fig:pipeline} upper left) for previous seen observation states. Intuitively, using auxiliary actions allows the policy model to better understand the multi-modal nature of expert action distributions. These pairs of point cloud and action constitute the supervised learning dataset $\mathcal{D}_{SL}$.

\subsubsection{Network Architecture}\label{sec:sl_network}
Our network takes a masked 3D point cloud as input. We adopt a ResUNet3D~\cite{resunet} and a lightweight Transformer~\cite{transformer} as backbone. We use a diffusion head to predict final action primitive parameters.
To facilitate the efficiency of training and inference with auxiliary actions, we design a special technique for parallel training and inference. 
We reorganize the data in self-attention layers of the Transformer to prevent information leakage between distinct action tokens. This allows our network to simultaneously take multiple auxiliary actions and diverse noise in parallel for each state during training. Furthermore, during inference, for each state, our model can output multiple ($N$) potential actions in parallel with only one pass. We use the following DDPM \cite{DDPM} loss function for supervised learning:
\vspace{-0.2cm}
\begin{equation}
    L_{SL} = \mathbb{E}_{(\mathbf{a}_0, \mathbf{P})\in\mathcal{D}_{SL}, t, \epsilon} \| \epsilon - \epsilon_\theta(\mathbf{a}_t, \mathbf{P}, t)\|_2^2
    \vspace{-0.1cm}.
\end{equation}

\subsection{Preference Learning by DPO Finetuning}
\vspace{-0.15cm}
\label{sec:pl}
To alleviate distribution shifts in long-horizon tasks, we collect a new round of on-policy data by running rollouts with the supervised model trained in Sec. \ref{sec:sl}. We annotate the data with human preferences, then use DPO~\cite{diffusion_dpo} to finetune the supervised model. 
Next, we will describe how we collect preference data and illustrate the learning algorithm.

\subsubsection{Data Collection}
When we run rollouts with the pre-trained supervised model, we will record $N$ predicted potential actions $\mathbf{A} = \{\mathbf{a}\}_1^N$ for each given state in one single pass.
Annotators will first annotate an optimal action $\mathbf{a}^0$ then do the comparisons between these $N$ predicted actions. Because $N$ may be large, we design an efficient ranking-based preference data annotation strategy (see Fig.~\ref{fig:pipeline} lower left).
During annotation, since some poor actions cannot be compared, annotators divide these actions into two groups: the better, rankable ones $\mathbf{A}_{r}$ and the poorer, unrankable ones $\mathbf{A}_{ur}$.
Then actions in $\mathbf{A}$ are sorted and the preference data are generated by performing the Cartesian product among or between these groups, which is expressed as
\vspace{-0.2cm}
{\small
\begin{equation}
    \{(\mathbf{a}^w, \mathbf{a}^l) | \mathbf{a}^w, \mathbf{a}^l\in \mathbf{A}^r, \mathbf{a}^w \succ \mathbf{a}^l\} 
    \cup 
    \mathbf{A}_{r}\times \mathbf{A}_{ur}
    \vspace{-0.2cm}
    \cup
    \{\mathbf{a}^0\} \times \mathbf{A}.
\end{equation}}Here, $\mathbf{a}^w\succ\mathbf{a}^l$ denotes action $\mathbf{a}^w$ win over action $\mathbf{a}^l$ in the ranking.
Through this, we construct the preference learning dataset $\mathcal{D}_{PL}$.
To enhance data efficiency, we prioritize using the more distant sample pairs in the sorted sequence during training.

\subsubsection{Learning Algorithm}
Once we have the preference dataset, we can finetune the policy model from a perspective similar to RLHF~\cite{christiano2017deep}.
The RLHF objective maximizes a reward model $r(\mathbf{a}, \mathbf{P})$ while regularizing the difference with the initial reference model.
Meanwhile, from the Bradley-Terry model~\cite{Bradley_Terry}, we also have another relation for $(\mathbf{a}^w, \mathbf{a}^l, \mathbf{P})\in \mathcal{D}_{PL}$, which is
\vspace{-0.2cm}
\begin{equation}
    \label{eq:Bradley_Terry}
    p(\mathbf{a}^w \succ \mathbf{a}^l|\mathbf{P}) = \sigma(r(\mathbf{a}^w, \mathbf{P}) - r(\mathbf{a}^l, \mathbf{P})).
    \vspace{-0.2cm}
\end{equation}
Following DPO~\cite{DPO}, we can indirectly train the RLHF objective by maximizing preference probability, bringing the policy closer to the optimal strategy under an implicit reward function.
When the policy is a diffusion model, Diffusion-DPO~\cite{diffusion_dpo} provides a corresponding loss function as
\vspace{-0.1cm}
{\small
\begin{align}
    L_{PL} &= -\mathbb{E}_{(\mathbf{a}^w_0, \mathbf{a}^l_0, \mathbf{P})\in \mathcal{D}_{PL}, t, \epsilon} \log\sigma  \nonumber \\ 
    &\{-\beta T [ 
    (\|\epsilon - \epsilon_\theta(\mathbf{a}^w_t, \mathbf{P}, t)\|_2^2 - \|\epsilon - \epsilon_{SL}(\mathbf{a}^w_t, \mathbf{P}, t)\|_2^2) -  \nonumber \\
    & (\|\epsilon - \epsilon_\theta(\mathbf{a}^l_t, \mathbf{P}, t)\|_2^2 - \|\epsilon - \epsilon_{SL}(\mathbf{a}^l_t, \mathbf{P}, t)\|_2^2)
    ]\}
\vspace{-0.2cm}
\end{align}}where $\beta$ is a regularization coefficient.
This objective can be intuitively seen as encouraging denoising to $\mathbf{a}^w_0$ and penalizing denoising to $\mathbf{a}^l_0$, while trying to keep the finetuned model's predictions close to the pre-trained model's.

\subsection{Parallel Inference with \textbf{R}eward-guided \textbf{A}ction \textbf{S}election}
\vspace{-0.15cm}
\label{sec:ras}
Preference learning enables models to differentiate between \textit{good} and \textit{bad} actions. However, with limited data, DPO finetuning can cause significant forgetting and performance degradation, a phenomenon observed in \cite{pal2024smaug}. 
To reduce abnormal actions and alleviate distribution shifts, we propose \textbf{R}eward-guided \textbf{A}ction \textbf{S}election (RAS) to choose from the multiple actions predicted by the supervised model trained in \ref{sec:sl} (see Fig.~\ref{fig:pipeline} right).

A key byproduct of DPO finetuning is the implicit reward function. We exploit this to ensure robust action selection during inference. For the 
$N$ potential actions predicted by the supervised model, we calculate the corresponding rewards and use a greedy strategy to select the action with the highest reward for execution.
This inference process can be formulated as
\vspace{-0.2cm}
\begin{equation}
    \hat{\mathbf{a}} = \mathop{\arg\max}_{\mathbf{a}\in \mathcal{M}(\mathbf{P})} r(\mathbf{a}, \mathbf{P})
    \vspace{-0.2cm}
\end{equation}
where $r$ is the reward function. As in Diffusion-DPO~\cite{diffusion_dpo}, we can compute $r$ as
\vspace{-0.2cm}
{\small
\begin{equation}
    r(\mathbf{a}_0, \mathbf{P})
    = -\mathbb{E}_{t, \epsilon}\beta T (\| \epsilon - \epsilon_{PL}(\mathbf{a}_t, \mathbf{P}, t)\|_2^2 - \| \epsilon - \epsilon_{SL}(\mathbf{a}_t, \mathbf{P}, t)\|_2^2)
    \vspace{-0.1cm}.
\end{equation}}It can be intuitively interpreted as evaluating the finetuned model's tendency of denoising to $\mathbf{a}_0$ while using the supervised model as a reference point.

To calculate rewards, we approximate the expectation through sampling. We observe that sampled values vary significantly across different diffusion timesteps $t$, with larger 
$t$ producing smaller values. Thus, we use only the smallest $10\%$ of timesteps for efficient reward calculation.

\textit{How to Understand  \textbf{R}eward-guided \textbf{A}ction \textbf{S}election (RAS)?}
In the context of generative probabilistic policy models, such as diffusion models, the predicted actions typically form a multimodal distribution, concentrating around several centroids. In our experiments, we observe that for previously unseen states, the optimal action is often included among the multiple predictions generated by the policy model. However, the relative probability of this optimal action is generally low, resulting in its infrequent sampling.
RAS can be understood as maintaining the original distribution of centroids while adjusting the assessment of their quality through reward-guided action selection.
When online data are limited, the discriminative quality prediction can be generalized more effectively and efficiently to unseen states.


\vspace{-0.1cm}
\section{Experiments}

\begin{figure*}[htbp]
  \centering
    \begin{minipage}{0.97\textwidth}
    \subfloat[]{
        \includegraphics[width=0.62\textwidth]{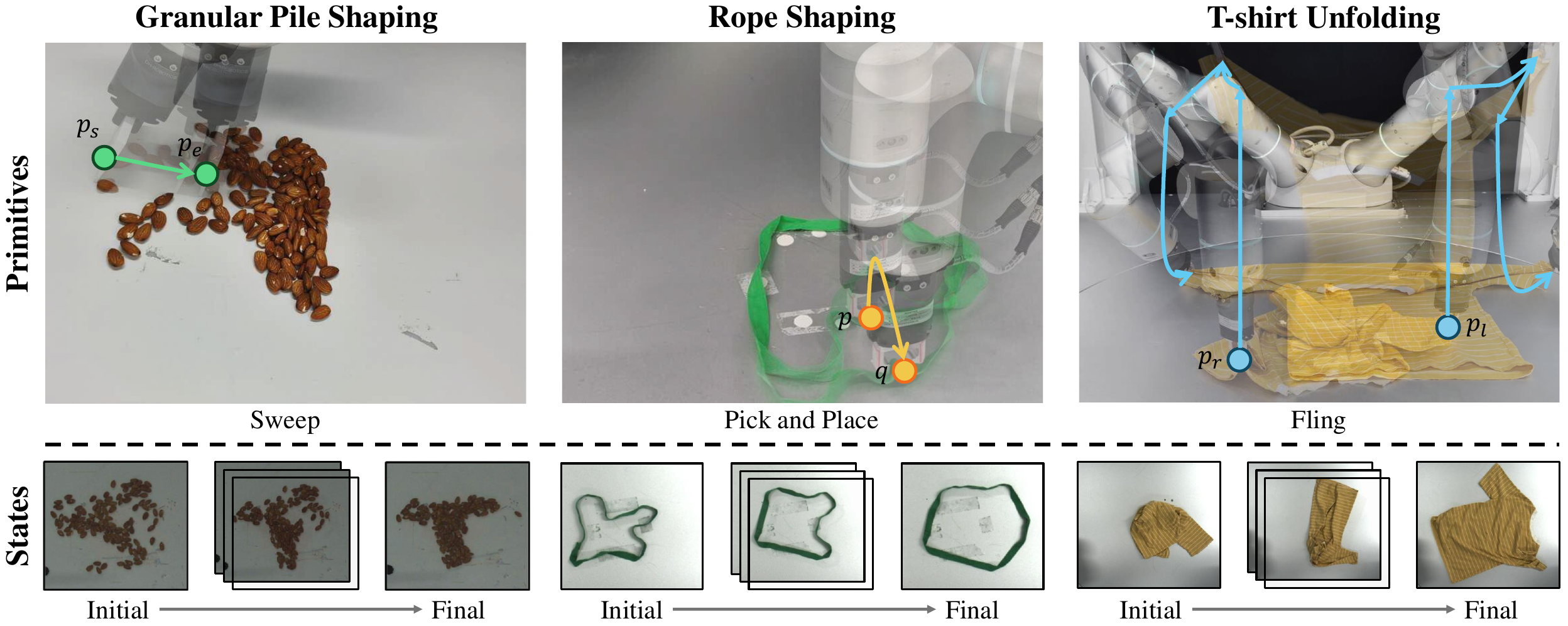}
        \label{fig:tasks_primitives}
    }
    \subfloat[]{
        \includegraphics[width=0.36\textwidth]{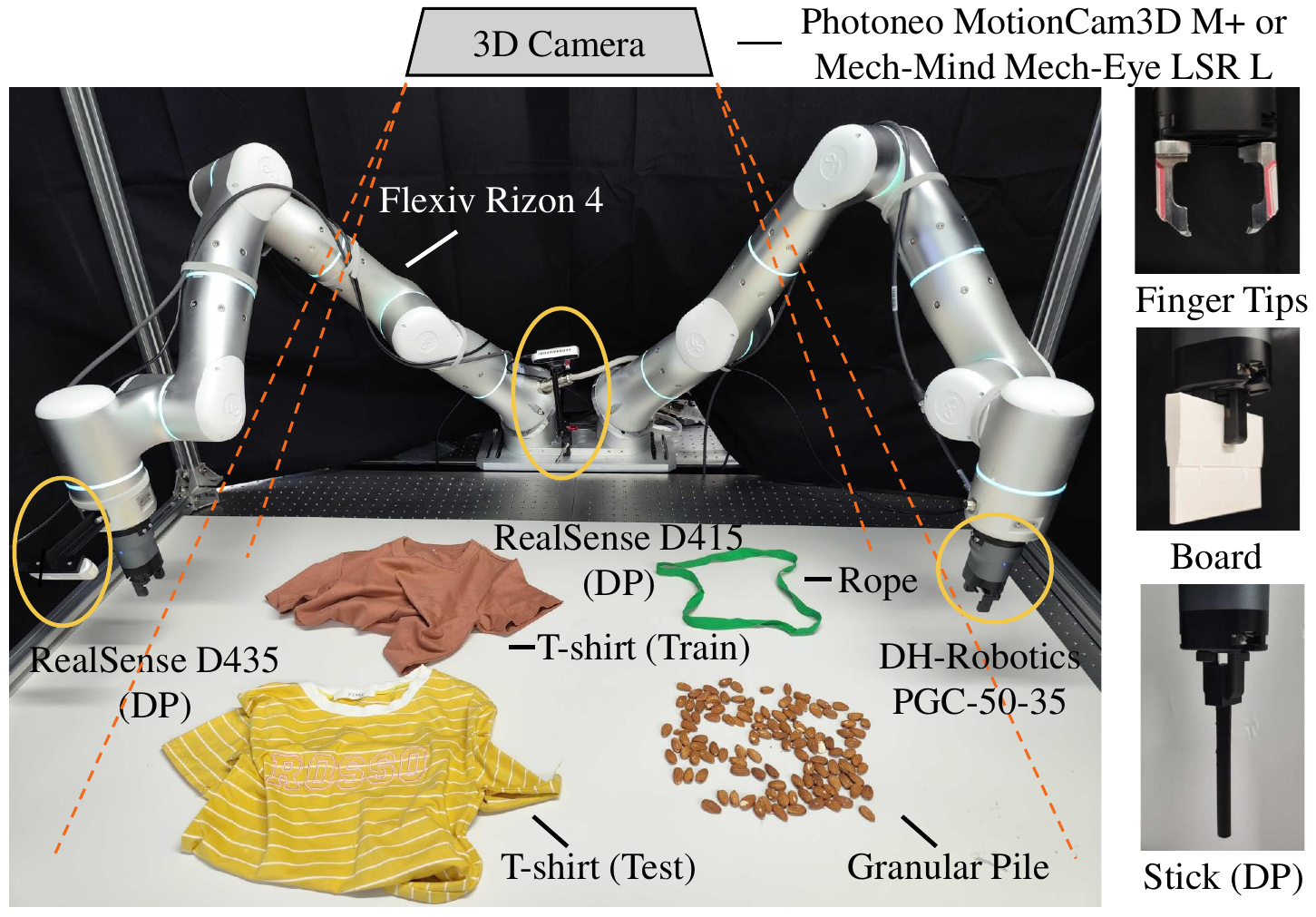}
        \label{fig:setup_tools}
    }
    \end{minipage}
  \vspace{-0.2cm}
  \caption{(a) Object states and primitives of each task. Beginning with a random complex state of an object, multiple steps of action primitives are performed to gradually achieve the target state. (b) Hardware setup and tools used in our real-world experiments. Devices and tools marked with DP are not used in primitive-based methods.}
  \label{fig:setup}
  \vspace{-0.5cm}
\end{figure*}
\vspace{-0.2cm}

We conduct experiments on three challenging real-world long-horizon manipulation tasks. We first describe the experimental design and baseline methods, then focus on examining \textbf{how does the model perform} and \textbf{what enables its capabilities} through quantitative and qualitative evaluations.

\subsection{Tasks and Hardware Setup}
\vspace{-0.15cm}
As shown in Fig.~\ref{fig:tasks_primitives}, we have designed three challenging long-horizon tasks: granular pile shaping, rope shaping and T-shirt unfolding.
These tasks involve 1D, 2D, and granular deformable objects and all start with complex initial states. 
Next, we describe the definition of each task.
\begin{itemize}
\item 
\textbf{Granular Pile Shaping}: In this task, the robot sweeps a disordered pile of granular objects (\ie nuts) into the shape of the character T.
We design a 3D-printed flat board as the robot tool and define the primitive parameters as $\mathbf{a} = (p_s, p_e)$, where $p_s$ and $p_s$ represent the start and end positions.
\item
\textbf{Rope Shaping}: In this task, the robot shapes a looped rope from a random shape into a circle using the pick-and-place primitive action $\mathbf{a} = (p, q)$, where $p$ and $q$ stand for the pick and place positions.
\item
\textbf{T-shirt Unfolding}: The goal of this task is to smooth out a short-sleeved T-shirt from a highly crumpled state. We use the fling action in Flingbot~\cite{ha2022flingbot} as the primitive $\mathbf{a} = (p_l, p_r)$, where $p_l$ and $p_r$ denote the left and right pick positions.
\end{itemize}

We employ intersection over union (IoU), coverage, and Earth Mover's Distance (EMD) calculated between the current state and the target state to evaluate the completion quality during the execution process. 

For hardware setup, the dual-arm platform and tools illustrated in Fig.~\ref{fig:setup_tools} are used to conduct all the experiments. 

\subsection{Baselines and Implementation Details}
\vspace{-0.15cm}
We design the following primitive-based methods for quantitative comparison.
\begin{itemize}
\item
\textbf{SL}: supervised model trained by offline data of stage 1.
\item
\textbf{SL + SL}: supervised model trained with offline data of stage 1 and the on-policy data (only the optimal actions) of stage 2.
\item
\textbf{DPO~\cite{DPO} + Implicit RAS}: DPO-finetuned model in stage 2 with implicit RAS during inference.
\item
\textbf{SL + Explicit RAS~\cite{Bradley_Terry}}: We implement an explicit reward model by adding a prediction head (similar to the action diffusion head) to the supervised pretrained network in stage 1. It is trained by directly optimizing Eq.~\ref{eq:Bradley_Terry} using preference data of stage 2. We use the pretrained supervised model in stage 1 for sampling actions and conduct reward-guided action selection (RAS) by explicit reward prediction.
\item
\textbf{SL + Implicit RAS \ie \framework~(Ours)}.
\end{itemize}
We train for 2000 epochs for supervised learning and 200 epochs for preference learning. All methods predict (sample) $N=8$ actions for each state during data collection and evaluation. We only capture object states before/after each action primitive for all primitive-based methods. We also implement Diffusion Policy (DP)~\cite{chi2023diffusionpolicy} with teleoperation data (RGB inputs, 10 FPS) as a primitive-free method only for qualitative comparison due to very different hardware and task settings. We annotate $K=9$ auxiliary actions for each state in the supervised dataset $D_{SL}$.
The specific dataset sizes are shown in the Tab. ~\ref{tab:dataset}.

\begin{table}[ht]
\vspace{-0.2cm}
\caption{
The dataset size for each task. PB and DP denote Primitive-Based methods and Diffusion Policy~\cite{chi2023diffusionpolicy}. \# seq. and \# states indicate the number of task sequences and states.
}
\vspace{-0.3cm}
\label{tab:dataset}
\begin{center}
\setlength{\tabcolsep}{1.5mm}{
\begin{tabular}{lcccccc}
\toprule
& \multicolumn{2}{c}{Granular Pile} & \multicolumn{2}{c}{Rope} & \multicolumn{2}{c}{T-shirt} \\
& \# seq.
& \# states
& \# seq.
& \# states
& \# seq.
& \# states
\\
\midrule
PB (Stage 1) &
$\sim$ 60 &
400 &
$\sim$ 30 &
200 &
$\sim$ 90 &
200
\\
PB (Stage 2) &
$\sim$ 25 &
200 &
$\sim$ 10 &
100 &
$\sim$ 50 & 
146
\\
\midrule
DP &
60
&
29807
&
50
&
9971
&
- &
-
\\
\bottomrule
\end{tabular}
}
\end{center}
\vspace{-0.7cm}
\end{table}

\subsection{Quantitative Evaluations}
\vspace{-0.15cm}

\begin{figure*}[htbp]
  \centering
    \subfloat[Granular Pile Shaping]{
        \begin{minipage}[c]{0.32\textwidth}
            \centering
            \includegraphics[width=\textwidth]{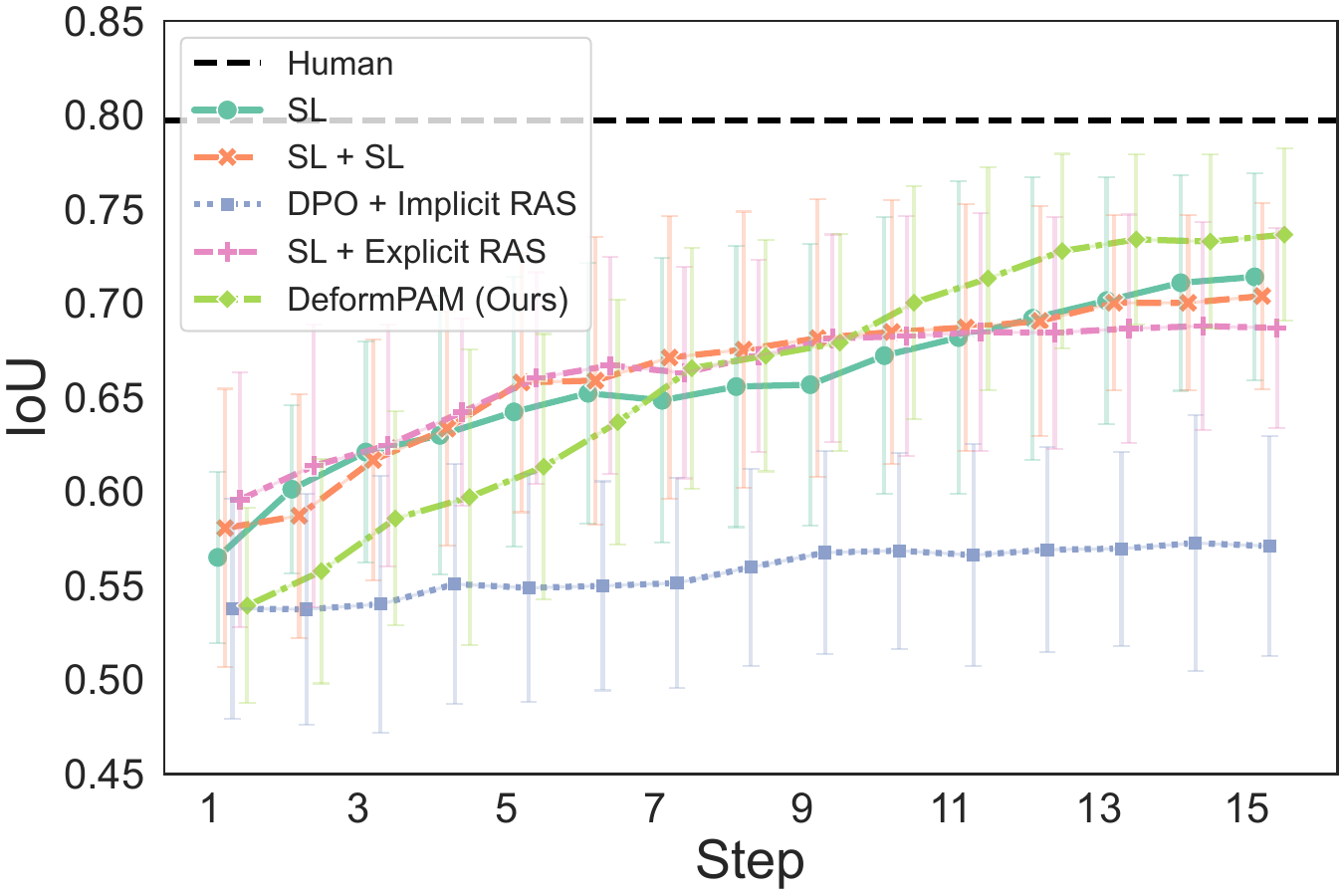} \\
            \includegraphics[width=\textwidth]{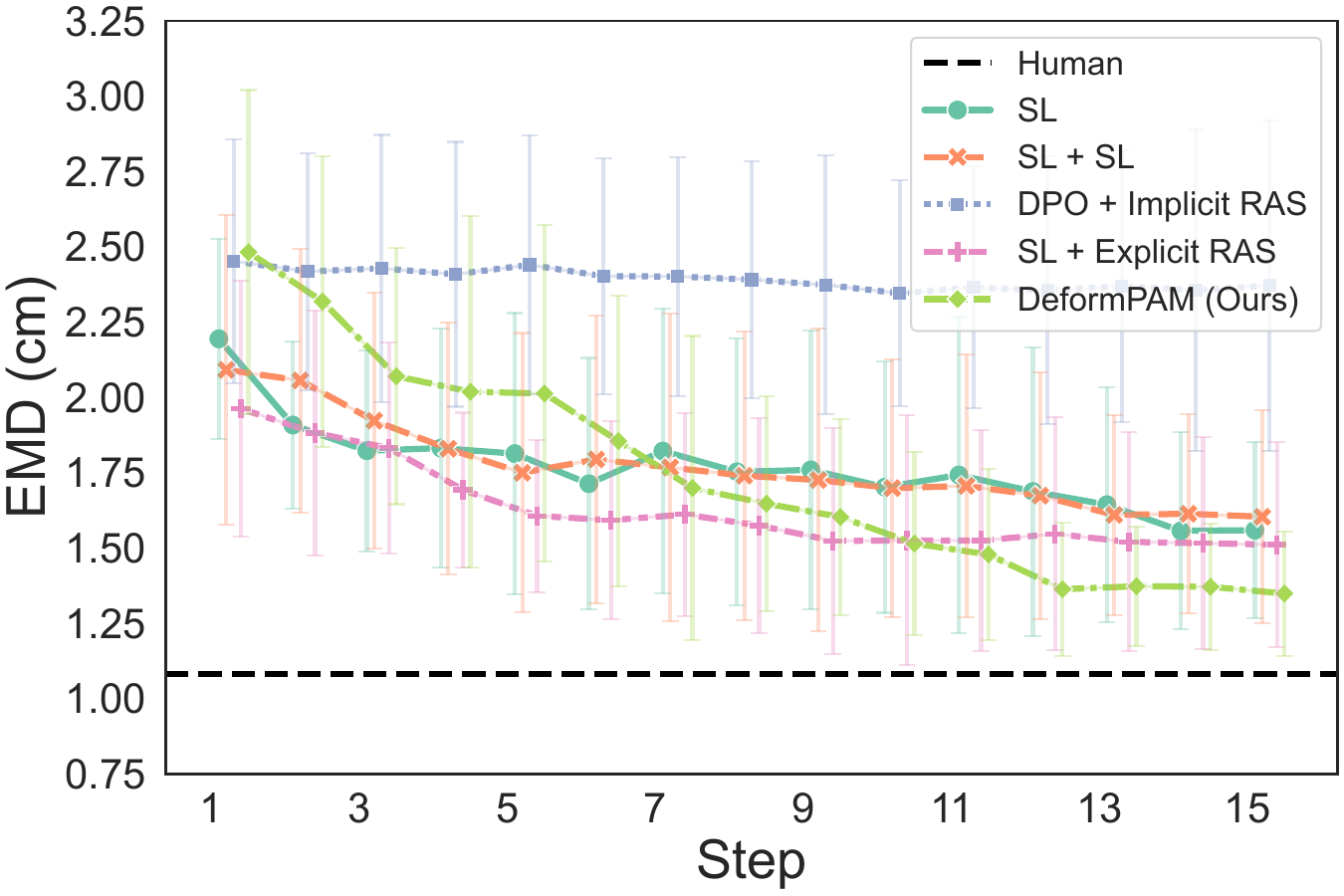}
        \end{minipage}
        \label{fig:metrics_nut}
    }
    \subfloat[Rope Shaping]{
        \begin{minipage}[c]{0.32\textwidth}
            \centering
            \includegraphics[width=\textwidth]{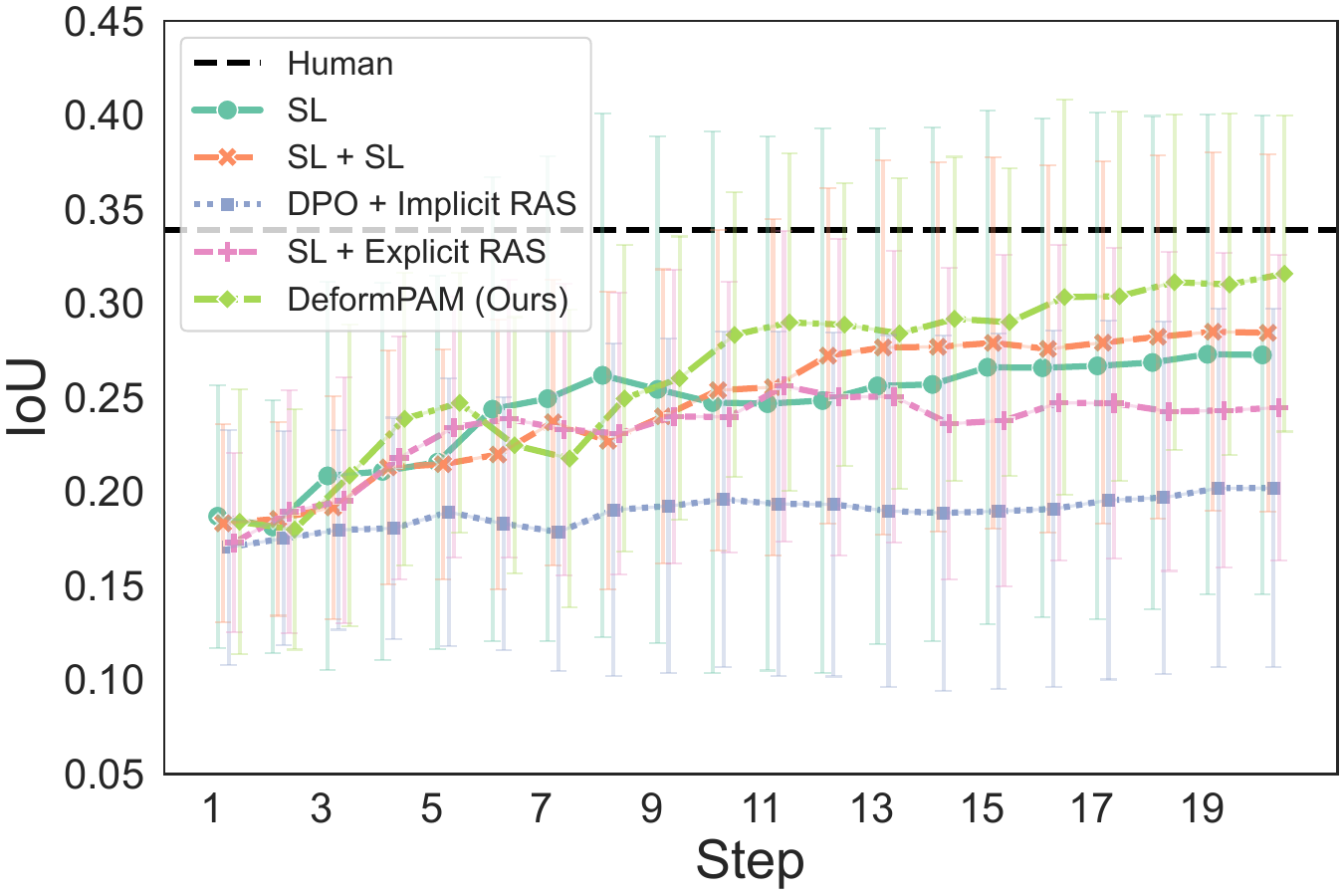} \\
            \includegraphics[width=\textwidth]{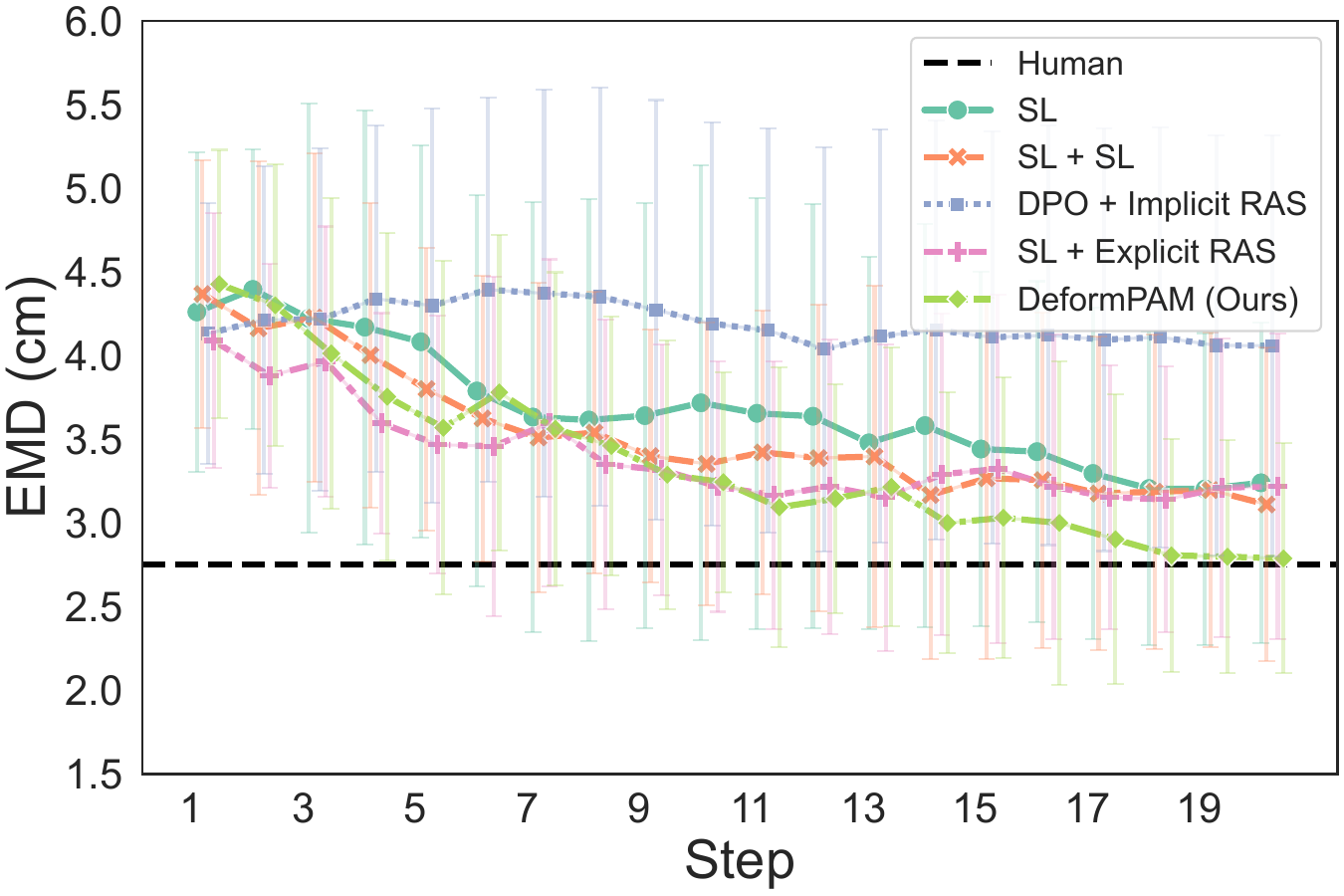}
        \end{minipage}
        \label{fig:metrics_rope}
    }
    \subfloat[T-shirt Unifolding]{
        \begin{minipage}[c]{0.32\textwidth}
            \centering
            \includegraphics[width=\textwidth]{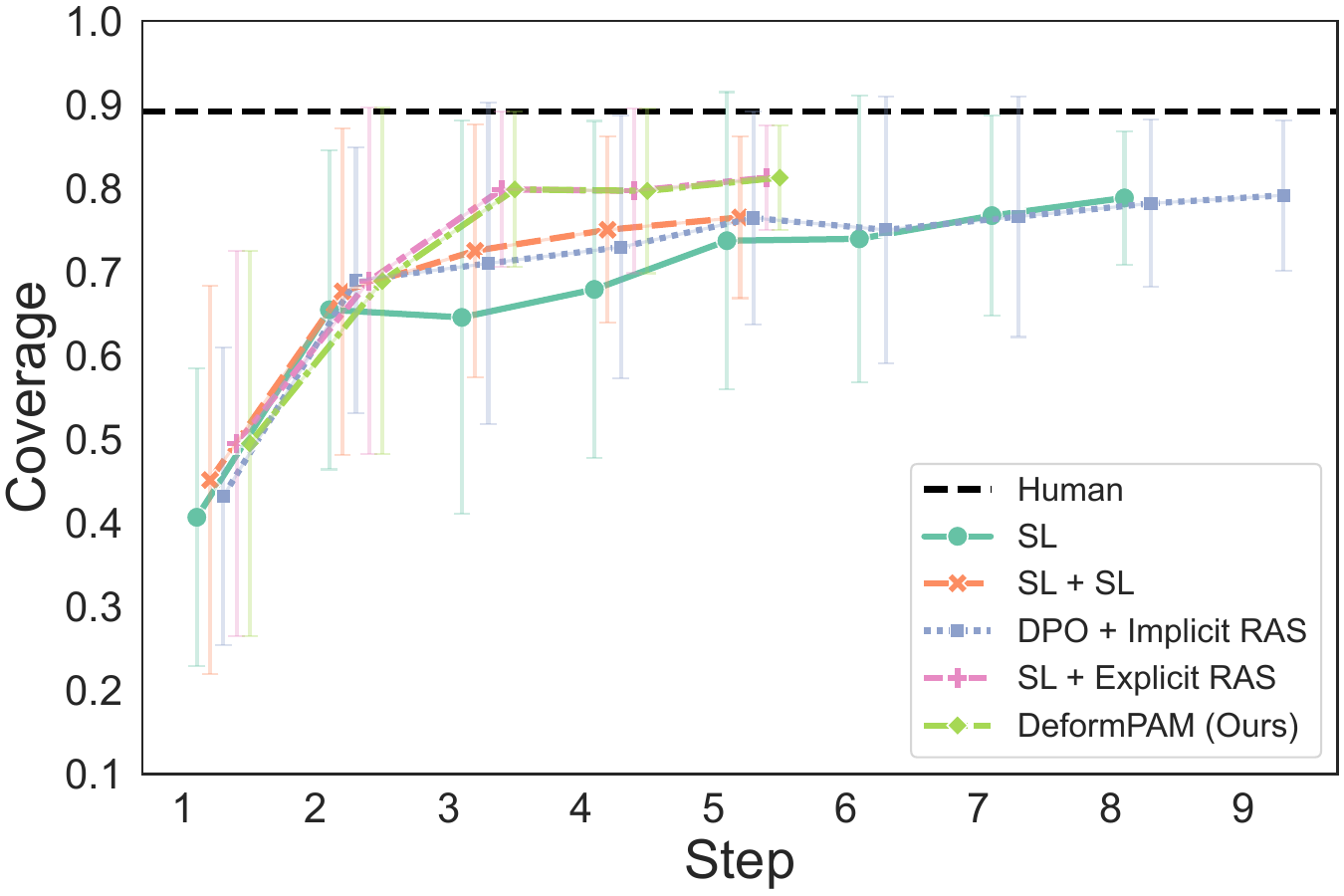} \\
            \includegraphics[width=\textwidth]{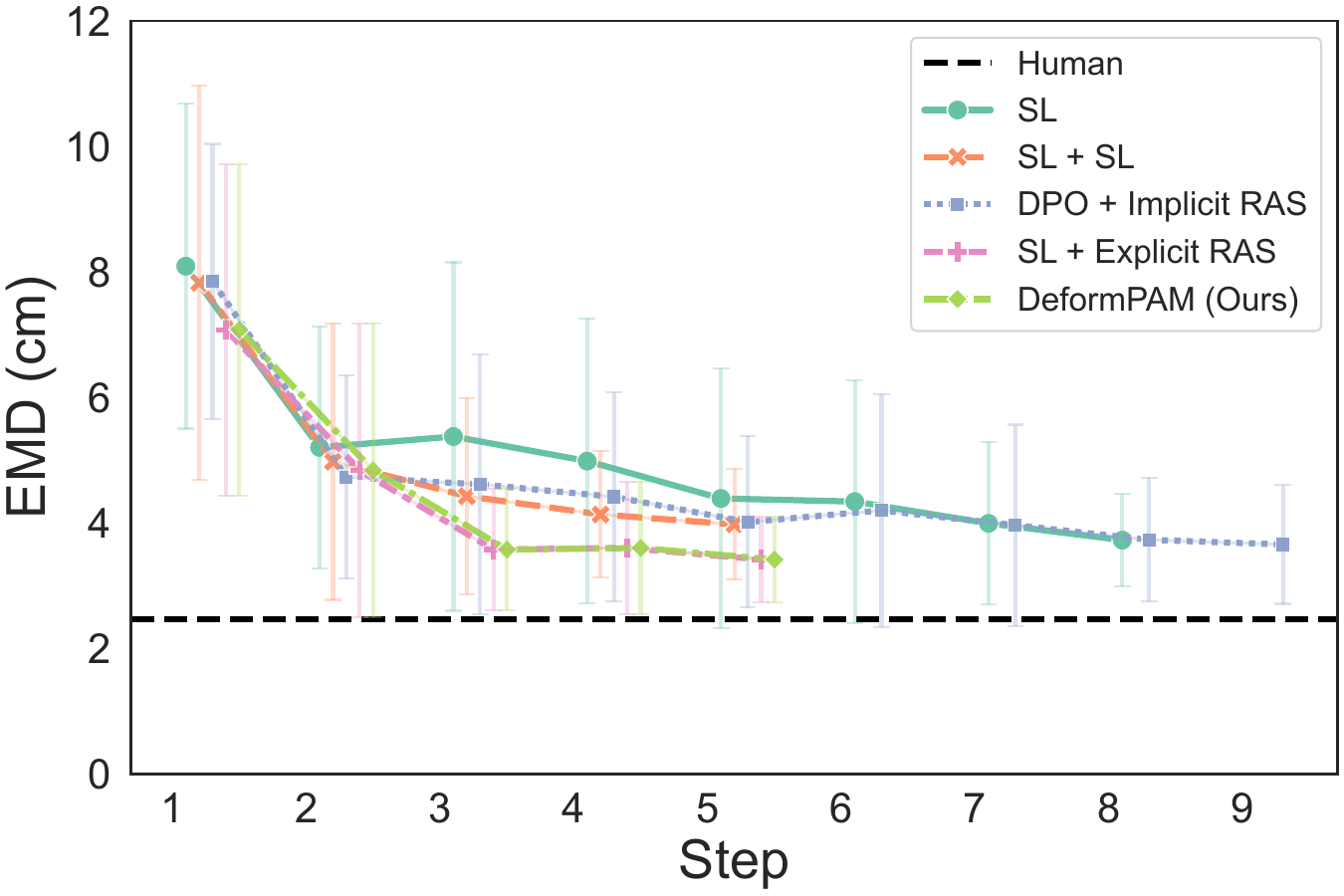}
        \end{minipage}
        \label{fig:metrics_tshirt}
    }
  \vspace{-0.2cm}
  \caption{Quality metrics per step on the three tasks. The results are calculated on 20 trials. Each trial ends until the policy already reaches its optimal state or exceeds the maximum steps. SL, DPO, RAS stand for the supervised model, DPO-finetuned model, and reward-guided action selection.}
  \label{fig:metrics}
  \vspace{-0.75cm}
\end{figure*}

The quantitative metrics from the real-world experiments are presented in Fig.~\ref{fig:metrics}.
We then illustrate the impacts of key components in our proposed framework and what enables its capabilities by answering the following questions. 

\textbf{Q1: Is using only supervised learning adequate for long-horizon tasks?}
As shown in Fig.~\ref{fig:metrics}, for the three tasks, with the help of reward-guided action selection, \framework~ leads to an increase in the final completion quality.
The variance in the quality metrics also tends to be smaller.
Meanwhile, SL is more likely to generate abnormal action and get trapped in an intermediate state, preventing further improvement in the quality curve.
The instability caused by these abnormal actions is mitigated through reward-guided action selection.

\textbf{Q2: How about training a supervised model with both off-policy and on-policy data?}
Training with on-policy data is another method to alleviate distribution shifts.
Although such a method can reduce the long-tail phenomenon of completion steps in Fig.~\ref{fig:metrics_tshirt}, the results in Fig.~\ref{fig:metrics_nut} and Fig.~\ref{fig:metrics_rope} indicate that SL + SL achieves only marginal improvements in harder tasks compared to the one using off-policy data.
Thus, employing reward-guided action selection is a more efficient method to enhance model performance.
\begin{figure}[htbp]
  \vspace{-0.15cm}
  \centering
  \subfloat[]{
      \includegraphics[width=0.23\textwidth]{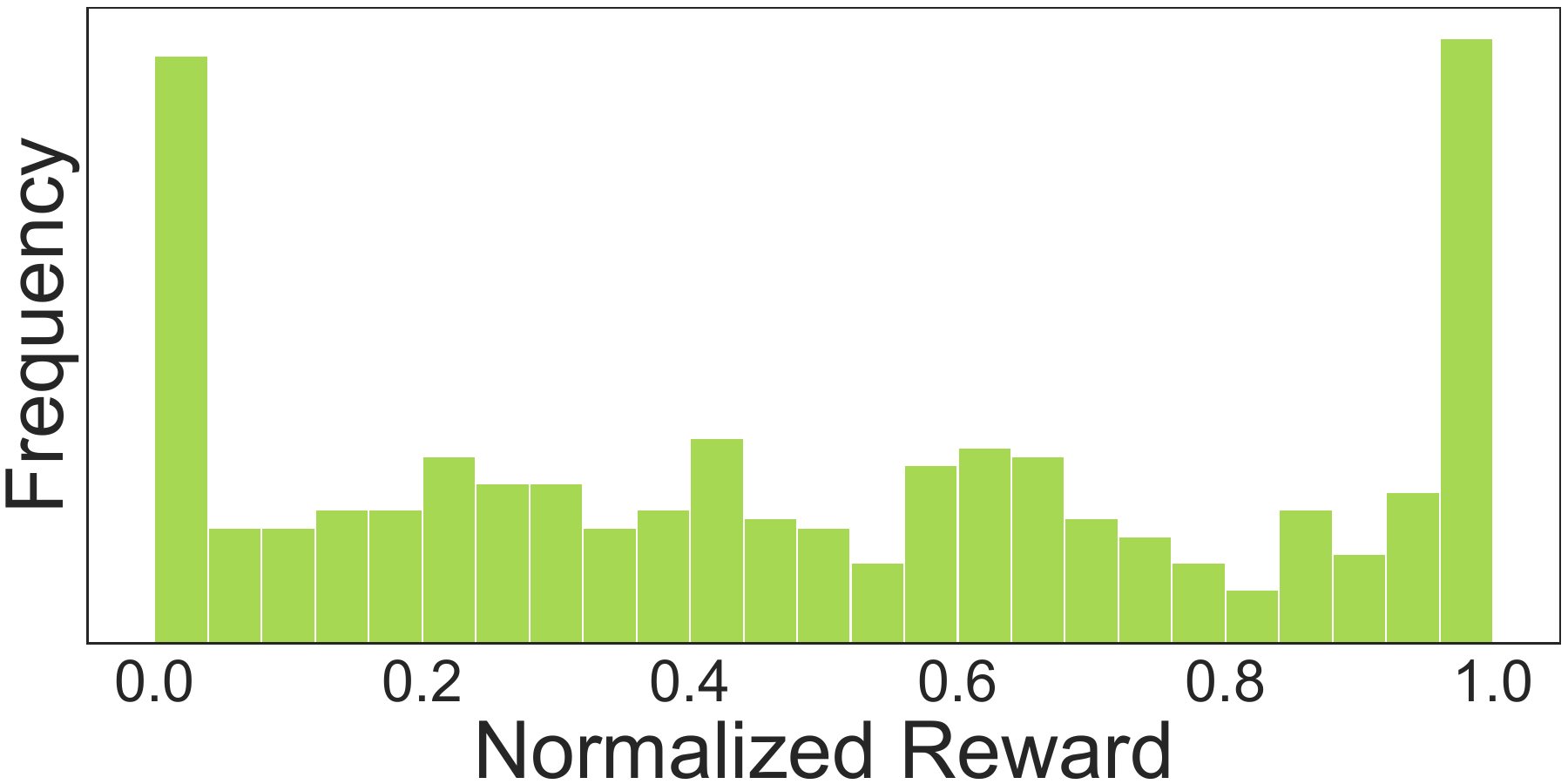}
      \label{fig:inference_reward}
  }
    \subfloat[]{
      \includegraphics[width=0.23\textwidth]{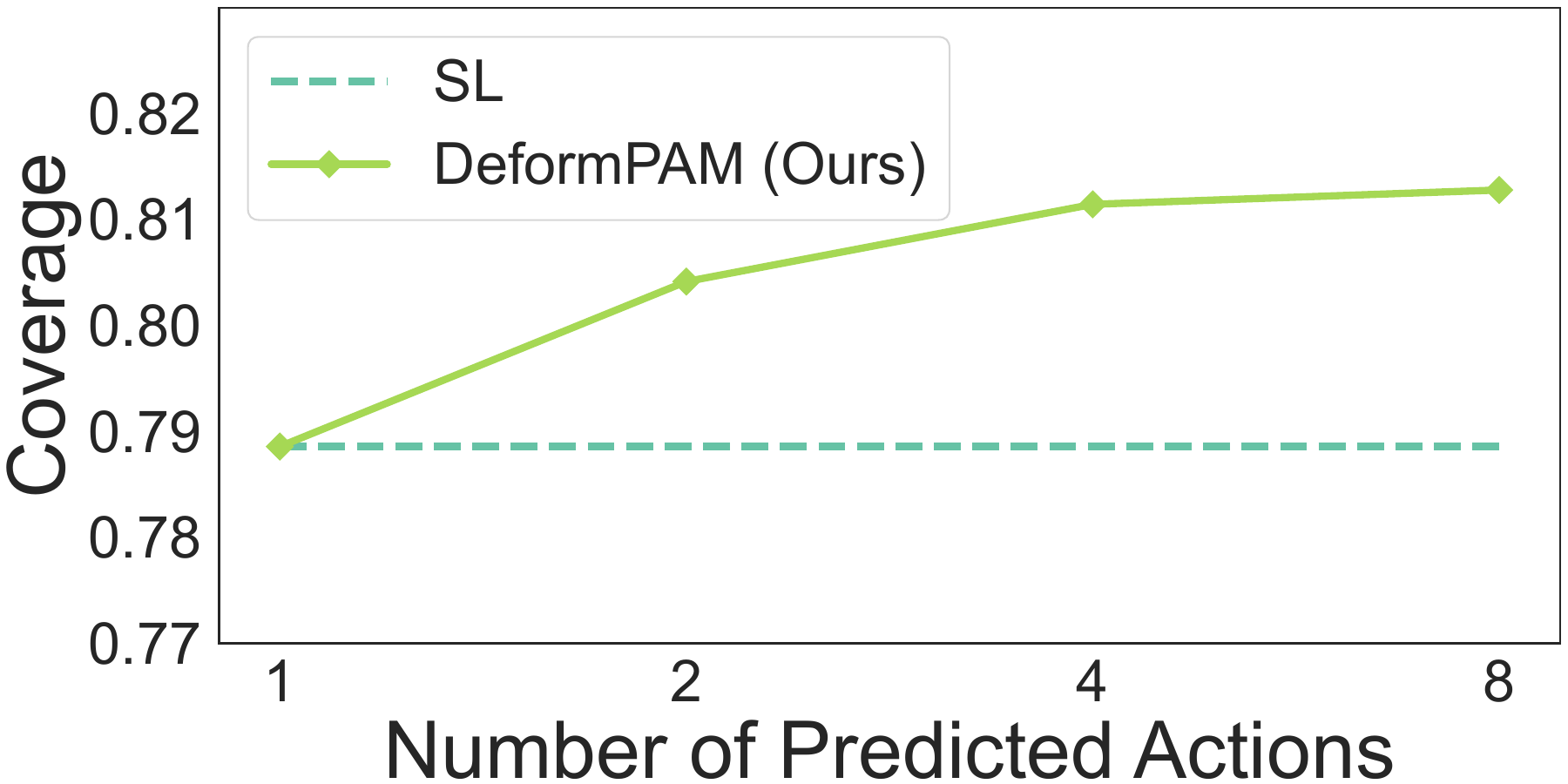}
      \label{fig:inference_coverage}
  }
  \vspace{-0.2cm}
  \caption{(a) Normalized reward distribution during inference when sampling $N=8$ actions. (b)  Average coverage for various numbers $N$ of predicted actions during inference. }
  \label{fig:inference}
  \vspace{-0.75cm}
\end{figure}

\textbf{Q3: Does employing the finetuned model to predict action primitives result in better performance?}
As seen in Fig.~\ref{fig:metrics_nut} and Fig.~\ref{fig:metrics_rope}, DPO + Implicit RAS performs worse on the shaping tasks compared to the standard \framework, and even underperforms the model using only supervised learning T-shirt Unfolding.
It is probably due to the forgetting issues~\cite{pal2024smaug} in DPO finetuning, which leads to worse action prediction quality.
This issue is more severe when data are limited, as is the case in this paper. 

\textbf{Q4: Is it more effective to extract the implicit reward model from DPO or to directly predict the reward?}
Besides extracting an implicit reward model, another way to obtain rewards is to directly train an explicit reward model with preference data.
From Fig.~\ref{fig:metrics_nut} and Fig.~\ref{fig:metrics_rope}, it can be found that for harder tasks like shaping, it is challenging for SL + Explicit RAS to achieve a high completion quality as the standard \framework.
This may be caused by reward overfitting when the size of the preference dataset is limited. In contrast, an implicit reward model from the DPO-finetuned model can fully leverage the action distribution learned during supervised learning.
This phenomenon is inconsistent with conclusions in NLP ~\cite{lin2024limited}, primarily because both pre-training and preference fine-tuning data are relatively abundant in NLP tasks.
Actually, as in Fig.~\ref{fig:metrics_tshirt}, an explicit reward model can also achieve a good performance in a simpler task (\ie T-shirt unfolding) with more data.

\textbf{Q5: How does reward-guided action selection (RAS) contribute to performance?}
We analyze the distribution of normalized implicit reward values during inference, as shown in Fig.~\ref{fig:inference_reward}.
This indicates that there is no positive correlation between the sampling probability of the action generation model and the predicted reward values, which suggests that employing RAS can serve as a quality reassessment.
From another perspective, we compare the performance between random sampling and reward-guided
action selection by adjusting the number $N$ of predicted actions during inference in the T-shirt unfolding task and computing the final coverage. As shown in Fig.~\ref{fig:inference_coverage}, as $N$ increases, the model's performance gradually improves.
This demonstrates that RAS enables the model to select superior samples, thereby benefiting from a greater number of samples.

\vspace{-0.3cm}
\subsection{Qualitative Results}
\begin{figure}[htbp]
\vspace{-0.3cm}
  \centering
    \subfloat[Granular Pile Shaping]{
        \hspace{-0.25cm}
        \includegraphics[width=0.097\textwidth]{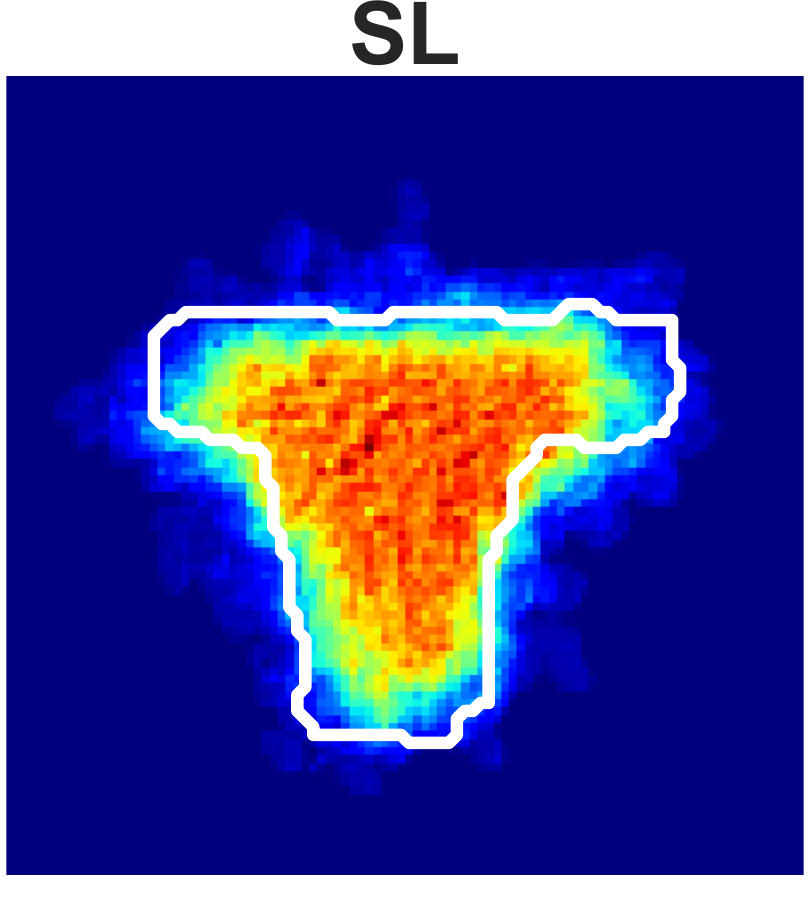}
        \hspace{-0.25cm}
        \includegraphics[width=0.097\textwidth]{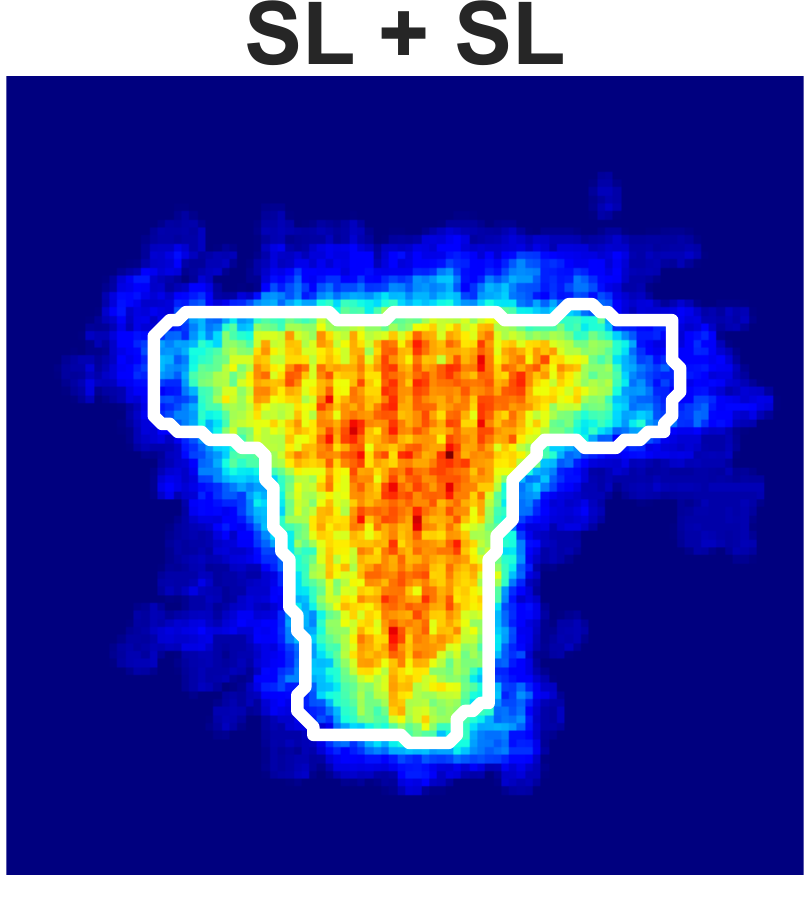}
        \hspace{-0.25cm}
        \includegraphics[width=0.097\textwidth]{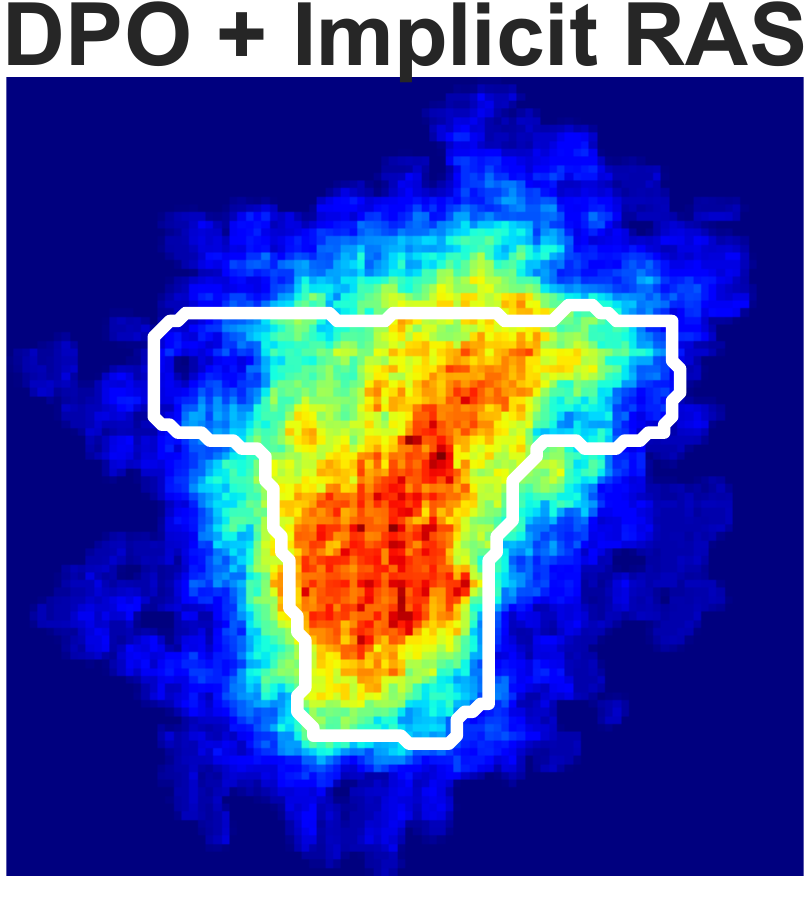}
        \hspace{-0.25cm}
        \includegraphics[width=0.097\textwidth]{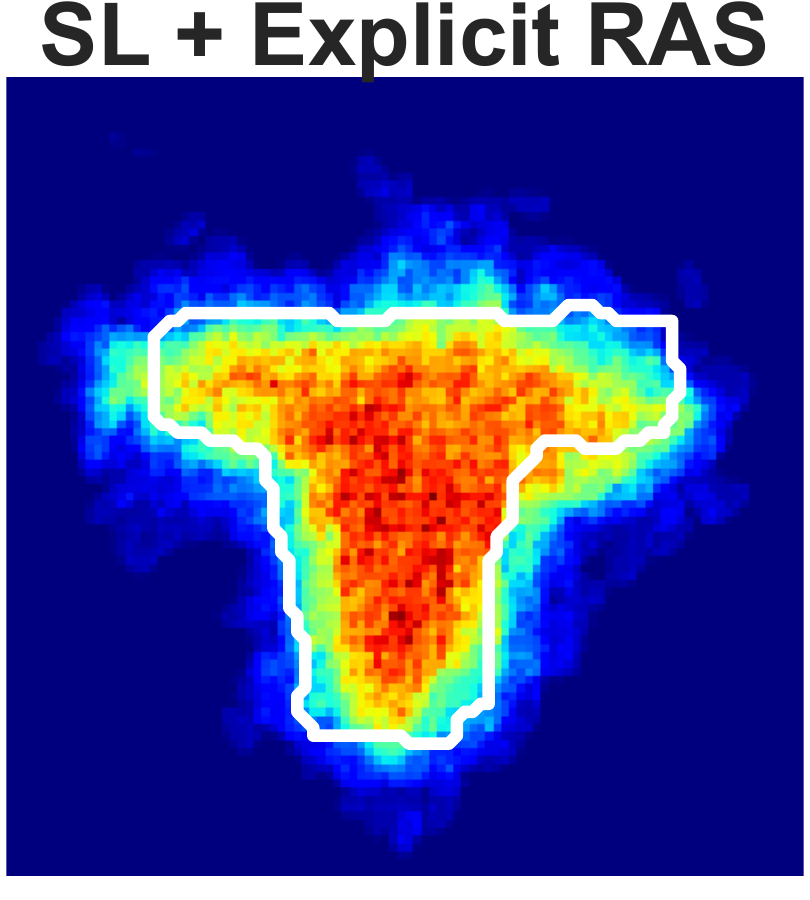}
        \hspace{-0.25cm}
        \includegraphics[width=0.097\textwidth]{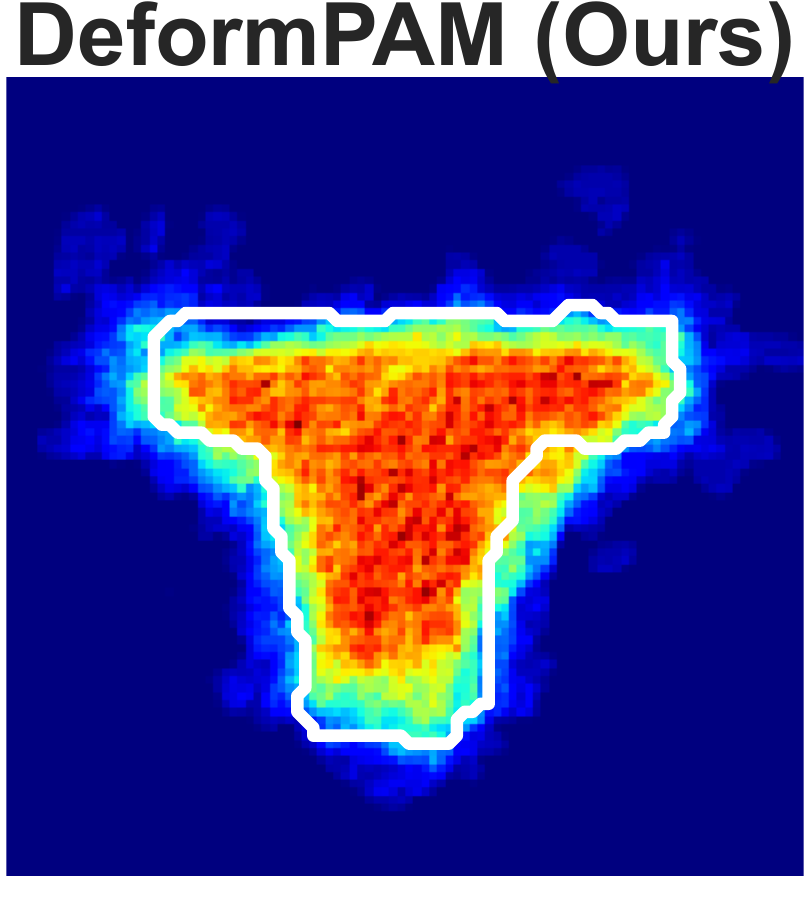}
    }\\
    \vspace{-0.3cm}
    \subfloat[Rope Shaping]{
        \hspace{-0.25cm}
        \includegraphics[width=0.097\textwidth]{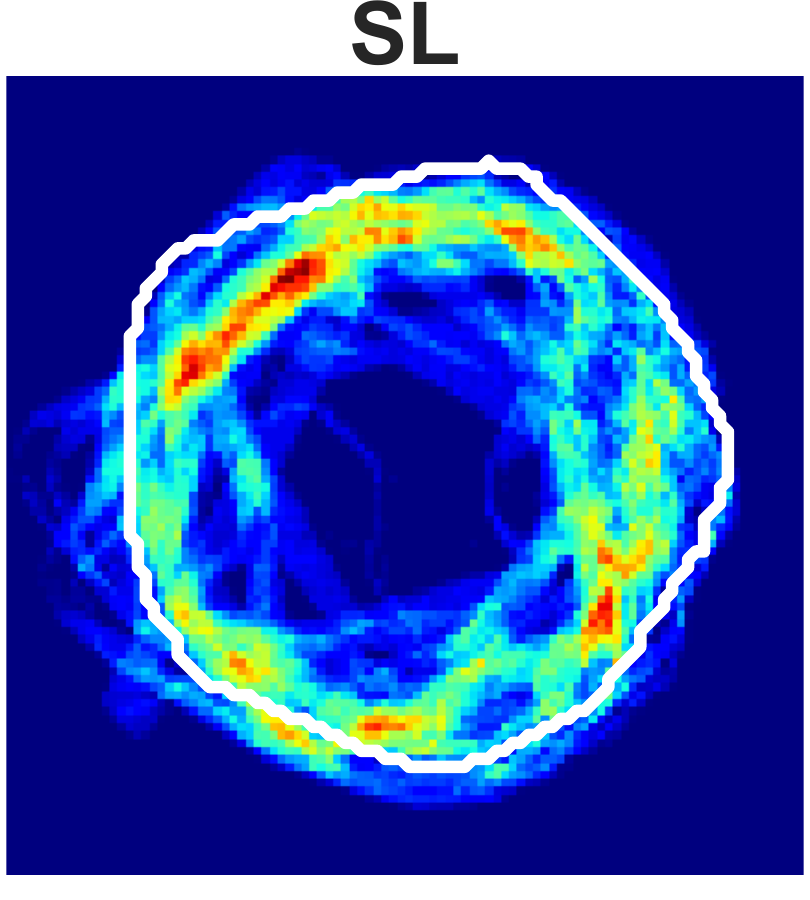}
        \hspace{-0.25cm}
        \includegraphics[width=0.097\textwidth]{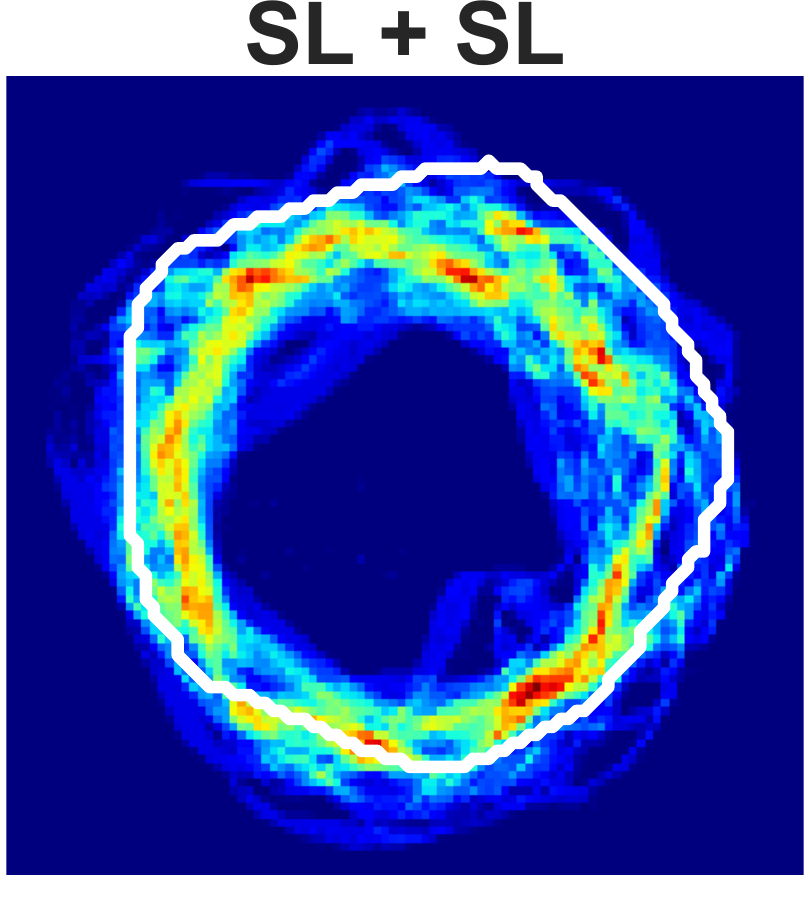}
        \hspace{-0.25cm}
        \includegraphics[width=0.097\textwidth]{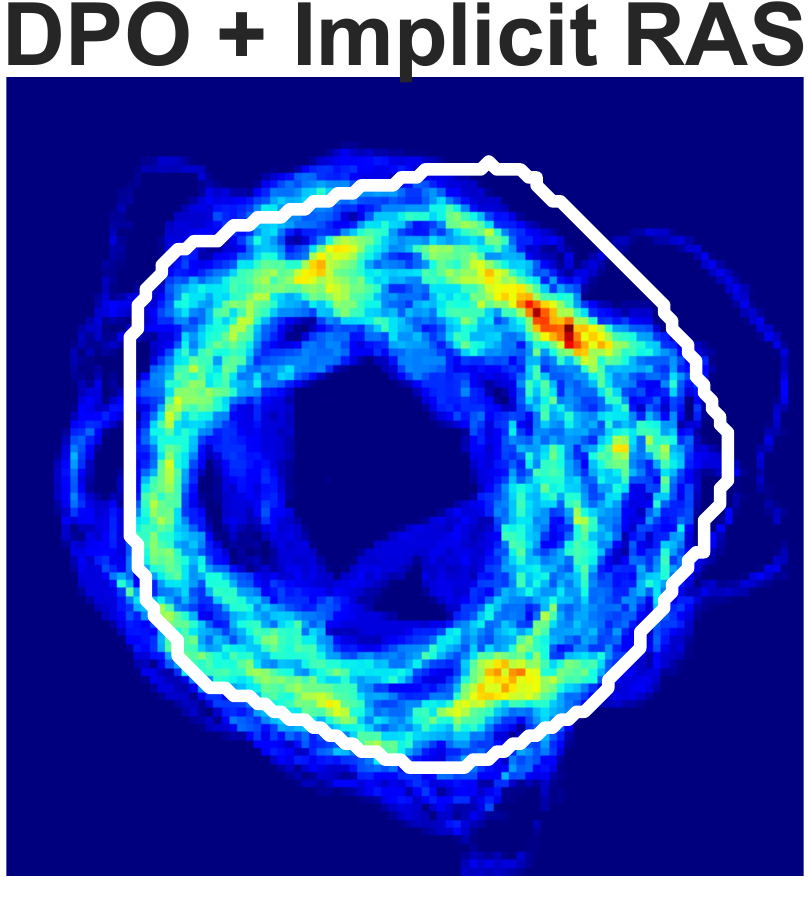}
        \hspace{-0.25cm}
        \includegraphics[width=0.097\textwidth]{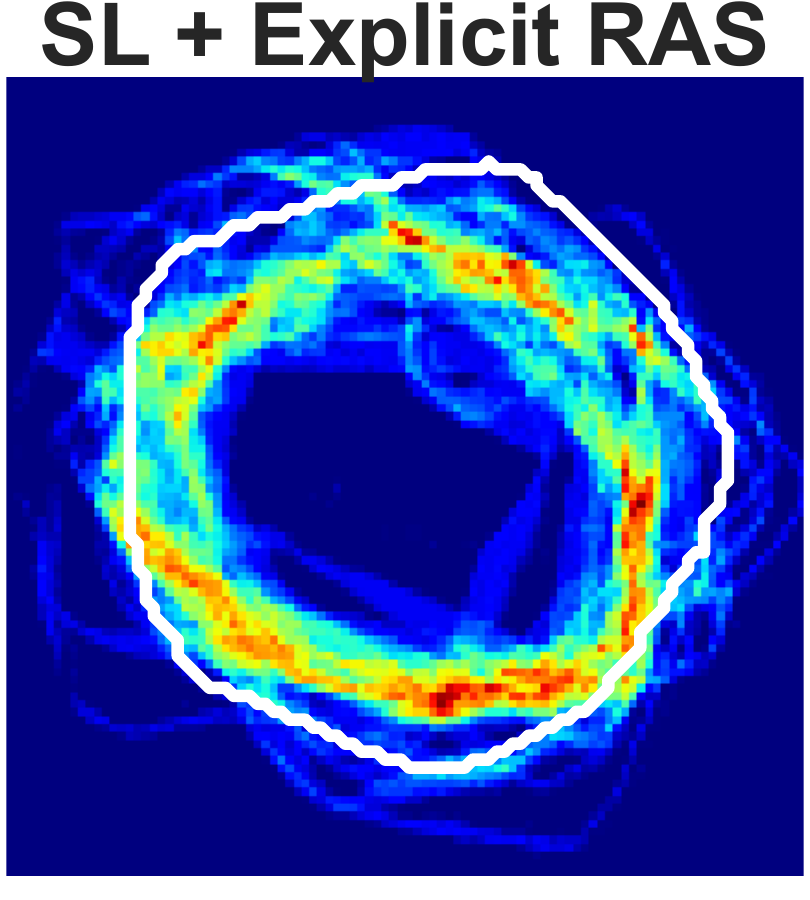}
        \hspace{-0.25cm}
        \includegraphics[width=0.097\textwidth]{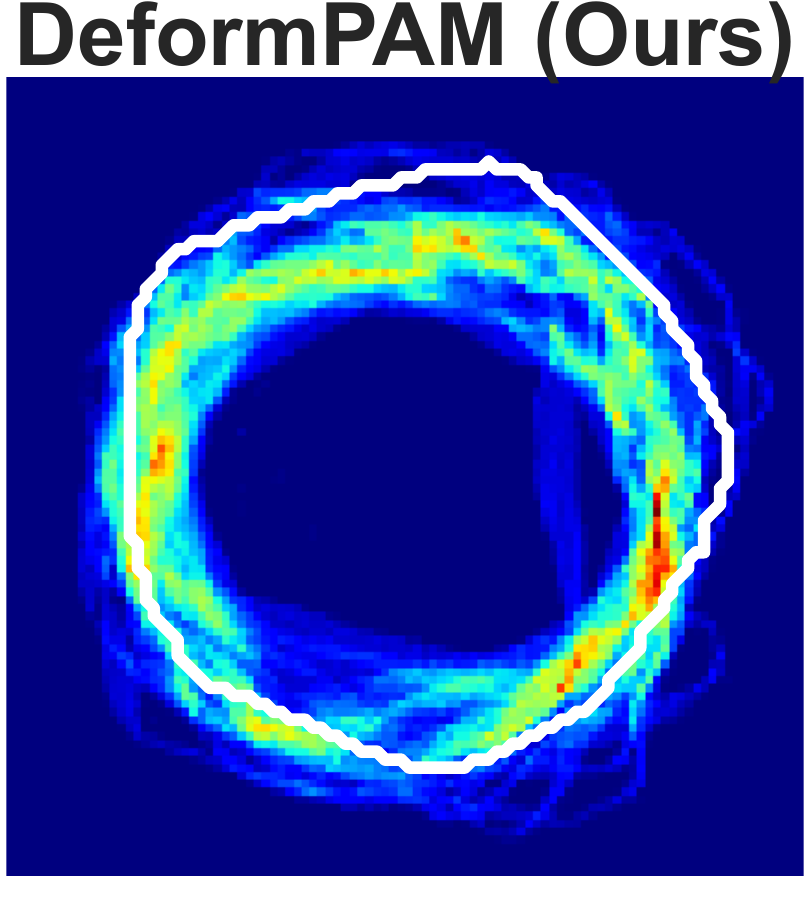}
    }
    \vspace{-0.2cm}
  \caption{Final-state heatmaps compared with the target states.}
  \label{fig:heatmap}
  \vspace{-0.5cm}
\end{figure}

\vspace{-0.15cm}
We draw the completion states of the granular pile shaping task and rope shaping task as heatmaps in Fig.~\ref{fig:heatmap}.
It is shown that our method achieves superior completion quality and exhibits lower variance. We also find that primitive-free methods like Diffusion Policy~\cite{chi2023diffusionpolicy} easily get stuck in unseen states with limited data. For more detailed results, please refer to the supplementary video and the \href{https://deform-pam.robotflow.ai}{project website}.


\vspace{-0.2cm}
\section{Conclusion}
\vspace{-0.2cm}
In this paper, we introduce DeformPAM, a novel framework for long-horizon deformable object manipulation that leverages preference-based action alignment to mitigate distributional shifts and enhance task performance. By integrating supervised learning with a preference learning model, DeformPAM employs reward-guided action selection to improve decision-making. Our experiments on three challenging real-world tasks demonstrate that DeformPAM enhances both task completion quality and efficiency compared to baseline methods. Future works could explore extending this approach to more complex tasks with multiple primitives.


\vspace{-0.2cm}
\section*{Acknowledgements}
\vspace{-0.2cm}
This work is supported by the Shanghai Committee of Science and Technology, China (Grant No. 24511103200) by the National Key Research and Development Project of China (No. 2022ZD0160102), Shanghai Artificial Intelligence Laboratory, XPLORER PRIZE grants.

{\small
\bibliographystyle{IEEEtran}
\bibliography{egbib}
}

\end{document}